\documentclass[preprint,12pt]{elsarticle}

\usepackage[a4paper, left=1in, right=1in, top=1in, bottom=1in]{geometry}

\usepackage[cmex10]{amsmath}
\usepackage{amssymb}
\usepackage{cases}
\usepackage{graphicx}
\usepackage{mathtools}
\usepackage{multirow}
\usepackage{booktabs}
\usepackage{color}
\usepackage{algorithm}
\usepackage{makecell}
\usepackage{algpseudocode, setspace}

\usepackage{changepage}

\usepackage{url}
\usepackage{float}
\usepackage{verbatim}
\usepackage{soul}
\usepackage{amsthm,amsmath,amssymb}
\usepackage{mathrsfs}

\soulregister\ref7 
\soulregister\cite7 

\usepackage[normalem]{ulem}

\allowdisplaybreaks

\begin{document}
	
	\begin{frontmatter}
		
		\title{M2WLLM: Multi-Modal Multi-Task Ultra-Short-term Wind Power Prediction Algorithm Based on Large Language Model}
		\tnotetext[1]{This work was supported by National Natural Science Foundation(7227139,U2268202) and Central Government Basic Research Business Fund for Universities(2024MS027)}
		
		\author[1]{Hang Fan}
		
		\author[2]{Mingxuan Li\corref{cor1}}
		\ead{lmx20@mails.tsinghua.edu.cn}

		\author[1]{Zuhan Zhang}
		
		\author[3]{Long Cheng}

            \author[4]{Yujian Ye}
		
		\author[1]{Dunnan Liu}
		
		\address[1]{School of Economics and Management, North China Electric Power University, 102206, Beijing, China}
		\address[2]{Department of Electrical Engineering, Tsinghua University, 100084, Beijing, China}
		\address[3]{School of Control and Computer Engineering, North China Electric Power University, 102206, Beijing, China}
            \address[4]{School of Electrical Engineering, Southeast University, 210094, Nanjing, China}
		
		\cortext[cor1]{Corresponding author}
		
		\begin{abstract} 
			The integration of wind energy into power grids necessitates accurate ultra-short-term wind power forecasting to ensure grid stability and optimize resource allocation. This study introduces M2WLLM, an innovative model that leverages the capabilities of Large Language Models (LLMs) for predicting wind power output at granular time intervals. M2WLLM overcomes the limitations of traditional and deep learning methods by seamlessly integrating textual information and temporal numerical data, significantly improving wind power forecasting accuracy through multi-modal data. Its architecture features a Prompt Embedder and a Data Embedder, enabling an effective fusion of textual prompts and numerical inputs within the LLMs framework. The Semantic Augmenter within the Data Embedder translates temporal data into a format that the LLMs can comprehend, enabling it to extract latent features and improve prediction accuracy. The empirical evaluations conducted on wind farm data from three Chinese provinces demonstrate that M2WLLM consistently outperforms existing methods, such as GPT4TS, across various datasets and prediction horizons. The results highlight LLMs' ability to enhance accuracy and robustness in ultra-short-term forecasting and showcase their strong few-shot learning capabilities.
		\end{abstract}
		
	\end{frontmatter}
	
	\begin{keyword}
		Wind Power Forecasting, Large Language Model, Multi-Modal Fusion, Few-Shot Learning
	\end{keyword}
	
	\section{Introduction}
	
	\subsection{Background}
	Wind energy, as a clean and renewable energy source, plays an increasingly important role in the transition of energy structures \cite{xue2015review}. The newly installed capacity of global onshore wind power is projected to exceed 100 GW by 2024 \cite{wang2024high}. However, the intermittency and uncertainty associated with large-scale wind power generation pose challenges to the stable operation of the power grid. Accurately predicting wind power output is crucial for system operation, resource scheduling, and improving the utilization efficiency of wind energy.
	
	During power system operation, forecasts of wind power output for forthcoming periods should be dynamically updated by leveraging historical time series data, such as past wind power output, wind speed, and other weather information. Depending on the temporal scope of the prediction, forecasting tasks can be categorized into ultra-short-term, short-term, and medium- to long-term forecasting \cite{hong2020energy}. Ultra-short-term forecasting, with a prediction horizon of minutes to a few hours ahead, is critical for real-time electricity markets. Short-term forecasting covers a horizon of several hours to a few days ahead and is used for bidding and scheduling in day-ahead electricity markets. Medium- to long-term forecasting is applied for long-term market planning and resource allocation.
	
	This paper focuses on ultra-short-term wind power forecasting, which aims to predict wind power for the subsequent minutes or hours based on historical time series data of power output and other pertinent auxiliary parameters.
	
	\subsection{Related Works}
	\subsubsection{Methods for Ultra-short-term Wind Power Forecast}
	Existing ultra-short-term wind power forecasting methods can be broadly categorized into two types: physical model-based methods and learning-based methods \cite{xue2015review}. Physical methods rely on wind power modeling based on wind turbine characteristics and wind speed. These methods predict future meteorological data, combine it with the topography surrounding the wind farm, calculate wind speed and direction at the hub height of the wind turbines, and ultimately derive the wind power output based on the turbine's power curve. 
	
	However, the accuracy of physical models is limited due to error accumulation during the modeling process. As a result, learning-based methods have become more widely used. These methods include machine learning approaches, deep learning techniques, and ensemble prediction methods, which are introduced as follows. 
	
	\textbf{(1) Machine learning-based methods} 
	
	Machine learning-based methods often employ bagging or boosting techniques for wind power regression. \cite{ARIMA_NP} combines ARIMA with random forest (RF), where RF is used to model the prediction residuals of ARIMA. In \cite{messner2019online}, online adaptive lasso estimation is integrated into vector autoregressive models for wind power prediction. By applying bagging or boosting technique, the model’s performance is improved. However, machine learning methods like RF are not well-suited for in-depth exploration of temporal relationships, limiting the ability to fully utilize sequence data. In \cite{XGBoost_BO}, eXtreme Gradient Boosting (XGBoost) is applied to short-term wind power forecasting, with Bayesian optimization used to fine-tune hyperparameters such as tree depth, learning rate, and the number of estimators. The results show significant performance improvement after hyperparameter optimization. \cite{hu2021hybrid} employs an automatic relevance determination algorithm to correct numerical weather prediction (NWP) data and integrates the corrected NWP with spatial correlations using Gaussian processes. In \cite{peng2023short}, a novel short-term wind power prediction model is introduced, leveraging stacked denoising autoencoders and multilevel transfer learning to address data scarcity and enhance prediction accuracy.
	
	\textbf{(2) Deep learning-based methods} 
	
	Deep learning-based methods achieve wind power forecasting using deep neural networks, such as recurrent neural networks (RNN) and long short-term memory networks (LSTM). In \cite{LSTM_EMD_kNN}, a bidirectional LSTM is used for wind speed prediction, and a novel empirical mode decomposition (EMD) method is proposed to address signal denoising. The proposed EMD method effectively handles measurement noise in the data, while the bidirectional LSTM structure captures more comprehensive information. However, the optimal design of model parameters still requires further study. In \cite{LSTM_WT_CSA}, the crow search algorithm (CSA) is employed to optimize the hyperparameters of the LSTM model for wind power forecasting. The learning rate, batch size, and input features are iteratively optimized within the algorithm. This method successfully eliminates irrelevant input features and achieves high accuracy with the optimized network. However, its performance needs to be validated on updated datasets, which may necessitate dynamic adjustments to the network structure. In \cite{LSTM_CNN_1} and \cite{ConvLSTM_2}, LSTM is combined with convolutional neural networks (CNN) for wind speed forecasting. The CNN effectively extracts spatial correlations in wind speed data, thereby fully leveraging the spatiotemporal information in wind power series. However, integrating CNN and LSTM results in a complex network structure, which requires a significant amount of data for effective training. To reduce the training burden, further exploration of methods to refine the network structure is necessary.
	
	With the rise of Transformer models, numerous recent studies have incorporated attention mechanisms into wind power forecasting models. The forecasting model in \cite{Attention_2} is inspired by the Transformer architecture, incorporating both encoder and decoder components along with their self-attention mechanisms. This model effectively captures long-range correlations in wind power data. However, it falls short in capturing the sequential relationships within the data—critical for wind power time series—which poses a significant challenge during model training. In \cite{Attention_1}, the forecasting model integrates LSTM with an attention mechanism. The encoder and decoder modules utilize historical wind power sequences and future weather forecast sequences, respectively, with LSTM extracting features from these sequential datasets. This customized network architecture fully leverages multi-source data and captures temporal information. However, it should be noted that this method relies on reliable information sources, as it depends heavily on comprehensive numerical weather forecast data. In \cite{LSTM_Attention_1} and \cite{LSTM_Attention_2}, the forecasting models adopt an LSTM-CNN-Attention architecture. LSTM and CNN first extract features from the time series data, and the attention mechanism then adaptively adjusts the weights of various temporal features. These architectures effectively utilize multivariate and multistep data for prediction. \cite{peng2023novel} proposes a cross-attention mechanism for CNN and BiLSTM to further enhance prediction accuracy.
	
	Graph Convolutional Networks (GCNs), as a deep learning method capable of effectively extracting spatiotemporal correlations, have been widely applied in ultra-short-term power forecasting for wind farm clusters. In \cite{fan2020m2gsnet}, a multi-modal, multi-task graph neural network is introduced for ultra-short-term power prediction in wind farm clusters. \cite{liang2024wpformer} employs a graph attention network to extract spatiotemporal relationships.
	
	\textbf{(3) Ensemble prediction methods}
	
	A single prediction model is rarely sufficient to handle the complexities of wind power processes. Therefore, the mainstream approach currently adopted is ensemble prediction. Many researchers also use traditional machine learning methods to divide wind processes into several spatiotemporal correlation patterns and design specific prediction models for each pattern to improve accuracy. In \cite{ye2022novel}, an innovative short-term wind power forecasting method is introduced, integrating frequency analysis, fluctuation clustering, and historical matching to enhance prediction accuracy by decomposing wind power sequences and matching historical data. \cite{von2021online} designs an online ensemble method to combine multiple forecasting models for probabilistic wind power prediction. In \cite{li2023adaptive}, an adaptive weighted combination forecasting approach is proposed, employing deep deterministic policy gradient (DDPG) to dynamically adjust model weights based on environmental changes. This enhances accuracy and adaptability compared to static combination models. \cite{liu2023spatio} and \cite{xu2023adaptive} propose an ultra-short-term wind farm cluster power forecasting model based on adaptive power fluctuation pattern recognition and optimal spatiotemporal graph neural network prediction. In \cite{zhao2024interpretable}, an interpretable ultra-short-term wind power forecasting method is presented, utilizing a multi-graph convolutional network with spatiotemporal attention and dynamic graph combination. \cite{yang2024short} predicts short-term wind power for farm clusters by employing a global information adaptive perception graph convolutional network, which features multiple characteristic graph structures. However, the complexity of the attention architecture requires a large amount of training data to mitigate the risk of overfitting. Finally, \cite{yang2024centralized} designs a short-term wind power prediction method using a dynamic graph neural network that adapts graph topology through node embedding to capture dynamic spatial correlations between wind farms. Additionally, it employs a decoupling error model to evaluate predictive performance.
	
	The essence of ensemble forecasting methods lies in accounting for the unique characteristics of different patterns when invoking prediction models. However, the complexity of wind processes makes it difficult to represent them with only a few pattern types. Current methods for mode division or clustering of wind processes carry a degree of subjectivity and often fail to adequately capture the unique features of each predictive scenario within the models. Furthermore, even the most advanced ensemble forecasting methods rely solely on numerical data, such as weather forecasts and historical power records, and do not fully utilize multi-modal information, such as textual descriptions.
	
	\subsubsection{Application of Large Language Model in Time Series Prediction}
	With the gradual rise of large language models (LLMs), many recent studies have started applying them to time series forecasting tasks. Pre-trained LLMs possess extensive knowledge reserves, enabling them to effectively learn and extract latent features from temporal data. \cite{GPT4TS} proposes a simple and unified framework that utilizes a frozen pre-trained GPT-2 model to achieve state-of-the-art performance in major time series analysis tasks, such as transportation and finance. Chronos \cite{ansari2024chronos} introduces a novel framework for pre-training probabilistic time series models, where time series data is tagged, and language models are trained on these tagged sequences using a cross-entropy loss function. In \cite{gruver2024large}, a time series forecasting model based on LLMs is designed and applied to financial, electricity load, and traffic forecasting scenarios. \cite{jin2023time} adapts large models for time series forecasting by employing reprogramming mechanisms and prompt engineering, achieving significant breakthroughs across numerous tasks. \cite{li2024urbangpt} draws on the successful experience of LLMs and proposes a spatiotemporal forecasting model named UrbanGPT. UrbanGPT combines a spatiotemporal dependency encoder with an instruction fine-tuning paradigm, enabling LLMs to understand complex spatiotemporal dependencies and produce more comprehensive and accurate predictions, even in data-scarce situations. In \cite{yu2023temporal}, financial text and transaction time series—two types of structured and unstructured data—are integrated into the large model framework for stock return prediction. LLMs can capture the complex relationships between these two types of data, while also eliminating the need for extensive text feature extraction required in previous studies, thus achieving better predictive results.
	
	The aforementioned studies integrate LLMs into various time series forecasting tasks across different fields using customized approaches, achieving promising results. However, research on LLM-based methods for renewable energy forecasting remains relatively limited, highlighting opportunities for further development in this area.

	\subsection{Research Gap and Contributions}
	
	Research on applying LLMs to wind power forecasting is still in its early stages. Studies like \cite{BERT4ST} and \cite{STELLM} have made valuable progress in this area. However, there remains an opportunity to further explore the alignment and integration of multi-modal information to better harness the potential of pre-trained LLMs. In this context, the model proposed in this paper fully leverages textual information in time series forecasting and effectively embeds numerical data into LLMs, thus maximizing their potential. The main contributions of this paper are as follows:
	
	\begin{itemize}
		\item \textbf{Multi-modal Capability}: We present a novel predictive framework that effectively combines textual information with temporal numerical data, thereby enhancing prediction accuracy. Unlike existing wind power forecasting methods that rely solely on temporal data, our approach integrates textual prompts and temporal data using a Prompt Embedder and a Data Embedder. This integration helps LLMs understand the predictive task and the distribution characteristics of the temporal data, enabling them to achieve higher prediction accuracy compared to methods that only use numerical data.
		\item \textbf{Leveraging Pre-trained Knowledge}: We introduce the Semantic Augmenter to translate temporal information into a format comprehensible to LLMs, allowing them to utilize their pre-trained knowledge to extract hidden temporal features. To adapt LLMs for wind power forecasting, we apply Low-Rank Adaptation (LoRA), fine-tuning only a small subset of parameters. This approach effectively leverages the pre-trained knowledge and domain adaptability of LLMs, improving prediction accuracy and enhancing their few-shot learning capability.
		\item \textbf{Extensive Experimental Validation}: We conduct a comprehensive evaluation of the proposed model alongside other state-of-the-art methods using historical wind power data from three different provinces and across various prediction horizons. The results demonstrate that leveraging LLMs for wind power forecasting improves both accuracy and robustness. Additionally, the LLM-based prediction model exhibits prominent few-shot learning ability, which is crucial when training data is insufficient for newly built wind farms.
	\end{itemize}

	\section{Problem Description and Preliminaries}
	\subsection{Problem Definition}
	Traditional ultra-short-term wind power forecasting aims to predict the wind power output $\mathbf{Y}$ for the next $ \tau_f $ time steps of historical wind power data $\mathbf{X}$ of the previous $ \tau_h $ time steps and NWP data $\mathbf{Z}$ of the future $ \tau_n $ time steps. A forecast model defines a mapping $f$ from the input historical wind power data $\mathbf{X}$ and future NWP data $\mathbf{Z}$ to the forecast values $\mathbf{\hat{Y}}$, expressed as 
	
	\begin{equation}
	\mathbf{\hat{Y}} = f(\mathbf{X},\mathbf{Z}, \mathbf{w})
	\end{equation}
	where the forecasted values are the future outputs of $ M $ wind power stations, denoted as $ \mathbf{\hat{Y}} = (\mathbf{\hat{y}_1}, \dots, \mathbf{\hat{y}_M})\in \mathbb{R}^{M \times \tau_f}$, where $ \mathbf{\hat{y}_i} = (\hat{y}_{i,t+1}, \dots, \hat{y}_{i,t+\tau_f}) $ represents the predicted wind power output of station $ i $ for time $ t $. The input historical wind power data with $ C $ stations is denoted as $ \mathbf{X} = (\mathbf{x_1}, \dots, \mathbf{x_C}) \in \mathbb{R}^{C \times \tau_h} $,  where $ \mathbf{x_i} = (x_{i,t-\tau_h}, \dots, x_{i,t}) $ represents the historical data for station $ i $. The NWP data such as wind speed, wind direction, temperature, and pressure can be described as $ \mathbf{Z} = (\mathbf{z_{11}},\mathbf{z_{12}}, \mathbf{z_{13}},\mathbf{z_{1n}}, \dots, \mathbf{z_{C*n}}) \in \mathbb{R}^{(C*n) \times \tau_h} $, where  $ \mathbf{z_{ij}} = (z_{i,j,t+1},\dots, z_{i,j,t+\tau_n}) $ is the  $ j$-th NWP feature of the station  $i$. Given $N$ training samples, the model is trained by adjusting its weights $\mathbf{w}$ to minimize the forecast error. The objective function for model training is formulated as follows:
	
	\begin{equation}
	\min_{\mathbf{w}} \frac{1}{N} \sum_{i=1}^N \|\mathbf{Y}_i - \mathbf{\hat{Y}}_i\|^2 
	\end{equation}
	where $\mathbf{Y}_i$ represents the actual values of wind power output for sample $i$, and $\mathbf{\hat{Y}}_i$ denotes the model's prediction for the same sample. Here, $\|\cdot\|^2$ denotes the squared norm, indicating that the squared error is computed across all relevant dimensions of the matrices. 
	
	However, the wind power process is complex, and a single model cannot handle all scenarios. Even if we divide the wind process into several modes and design different prediction algorithms for each mode, the mode division cannot fully capture the complexity of the wind process. Therefore, we introduce a multi-modal, multi-task prediction model based on LLMs, which incorporates verbal descriptions of task information, historical wind power data, and weather information. Consequently, the forecast model can be reformulated as:
	
	\begin{equation}
	\mathbf{\hat{Y}} = f(\mathbf{X},\mathbf{Z}, \mathbf{V_1}, \mathbf{V_2},\mathbf{V_3},\mathbf{w})
	\end{equation}
	where $ \mathbf{V_1} $,$ \mathbf{V_2}$,$ \mathbf{V_3} $ are task information, historical wind power information, and weather information respectively. By employing such a method, we can utilize multi-modal text information related to wind farm power prediction tasks, and implement a model for multi-task prediction across multiple stations. The corresponding model design will be presented in Section \ref{section: model}.

	\subsection{Mechanism of a pre-trained LLMs} \label{subsection: Arch_LLM}
	\begin{figure}[htbp]
		\centering
		\includegraphics[width=0.8\linewidth]{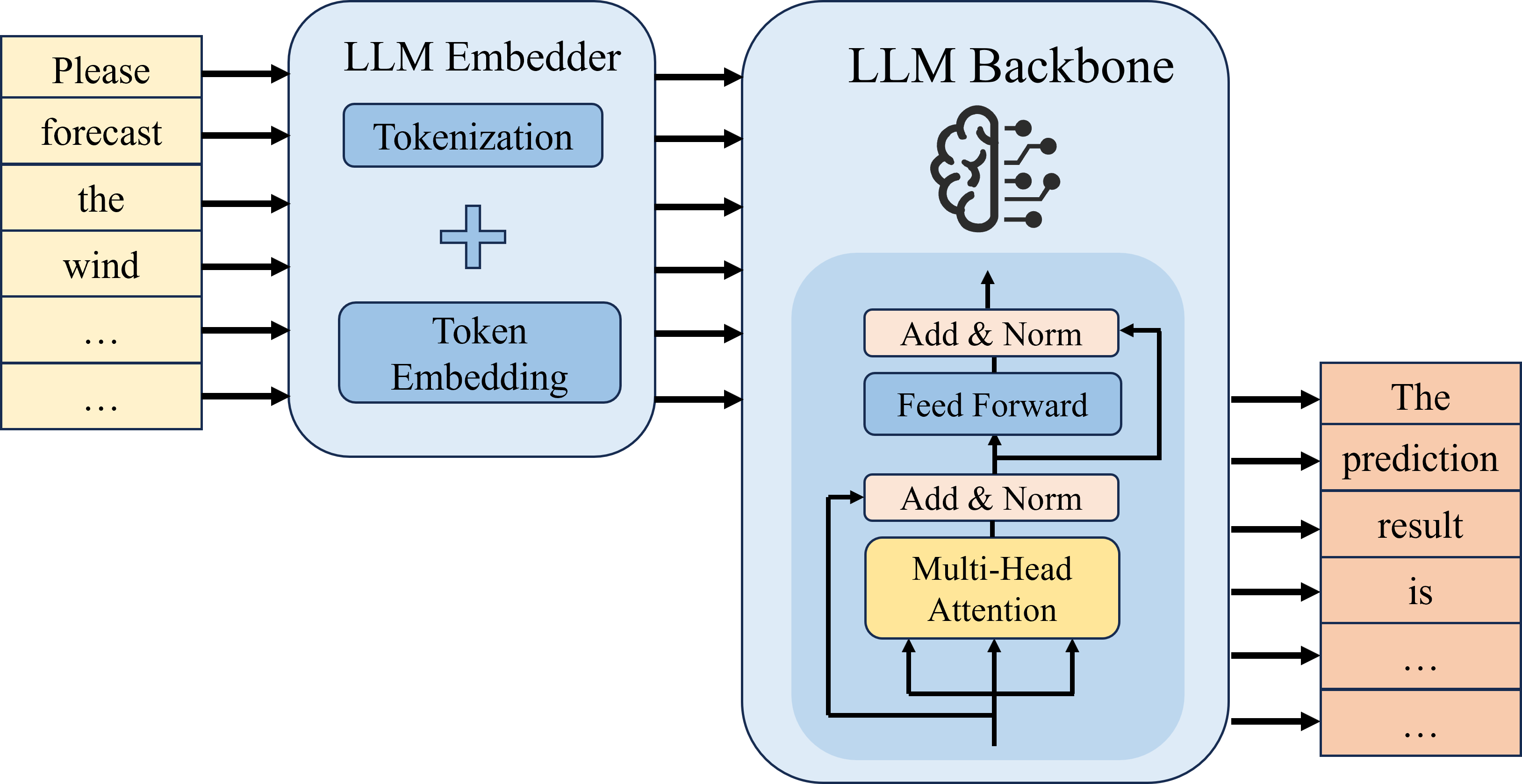}
		\caption{Demonstration of a pre-trained LLM utilizing token embedders and backbone models for text generation tasks}
		\label{fig:LLM_example}
	\end{figure}

	Before introducing the proposed method, we first introduce the mechanism of LLMs, and highlight the challenges when applying LLMs directly to wind power forecasting tasks.
	
	The mechanism of LLMs is illustrated in Fig. \ref{fig:LLM_example}. The input text is first tokenized by the LLMs, producing a series of word tokens. Each token is then mapped to a high-dimensional vector through the LLM's pre-trained word embedder (i.e., the token embedding process), which serves as input to the backbone of the LLMs. The LLM backbone is a decoder architecture composed of multiple layers of stacked Attention mechanisms. Leveraging its rich pre-trained knowledge and inherent text-processing capabilities, the LLMs generate coherent output text sequences.
	
	From this architecture, it is clear that pre-trained LLMs excel at processing textual data. But they have limited capabilities when it comes to understanding numerical data. If the input sequence contains numerical values, the LLMs still tokenize them, but cannot fully utilize the inherent information of these numbers. However, the input for wind power forecasting consists largely of time series data, which is predominantly numerical. Therefore, a key challenge is how to design an effective model that builds upon LLMs to maximize their potential for wind power forecasting.
	
	This challenge can be summarized into two key aspects: first, how to effectively integrate time series data with relevant textual prompts to leverage the LLM's strength in text understanding, and thereby gain deeper insights into wind power forecasting tasks; and second, how to efficiently feed time series data into the LLMs, so that it can better comprehend such inputs. The following section addresses these two issues by proposing an LLM-based architecture for wind power forecasting.

	\section{Methodology} \label{section: model}
	
	\subsection{Model architecture}
	
	To resolve the challenges mentioned in \ref{subsection: Arch_LLM}, the following modules are considered in this
	\begin{figure*}[htbp]
		\centering
		\includegraphics[width=\textwidth]{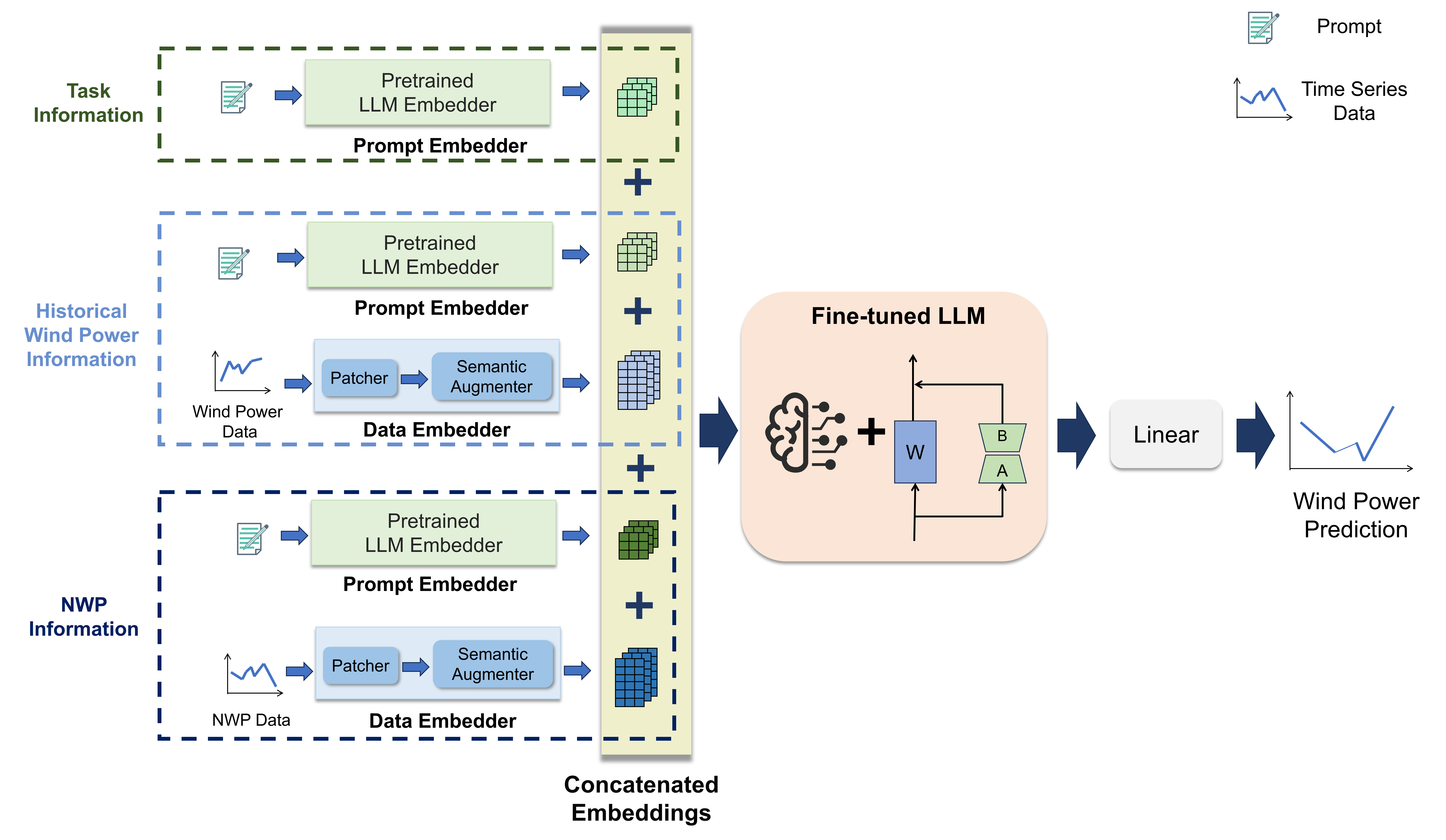}
		\caption{Overall architecture of the proposed model integrating diverse prompts and data}
		\label{fig:Flowchart}
	\end{figure*}
	 paper. Firstly, the prompt for the wind forecast task is designed. Secondly, the input time series data are patched and augmented with pre-trained word embeddings using a cross-attention mechanism, which enhances the semantic information and becomes a reasonable input of LLMs.  Thirdly, the LLM model is fined-tuned using limited parameters using LoRA. Based on these procedures, the overall architecture of the proposed model is depicted in Fig. \ref{fig:Flowchart}.
	
	Based on this structure, the proposed model effectively integrates textual and numerical data, leveraging the pre-trained knowledge of the LLMs to a greater extent, allowing it to better adapt to the wind power forecasting task. The following sections will provide a detailed description of the key components of this model, focusing on prompt design, numerical data embedding, and fine-tuning of LLMs.
	
	\subsection{Prompt Embedder}
\begin{figure}[htbp]
		\centering
		\includegraphics[width=0.7\linewidth]{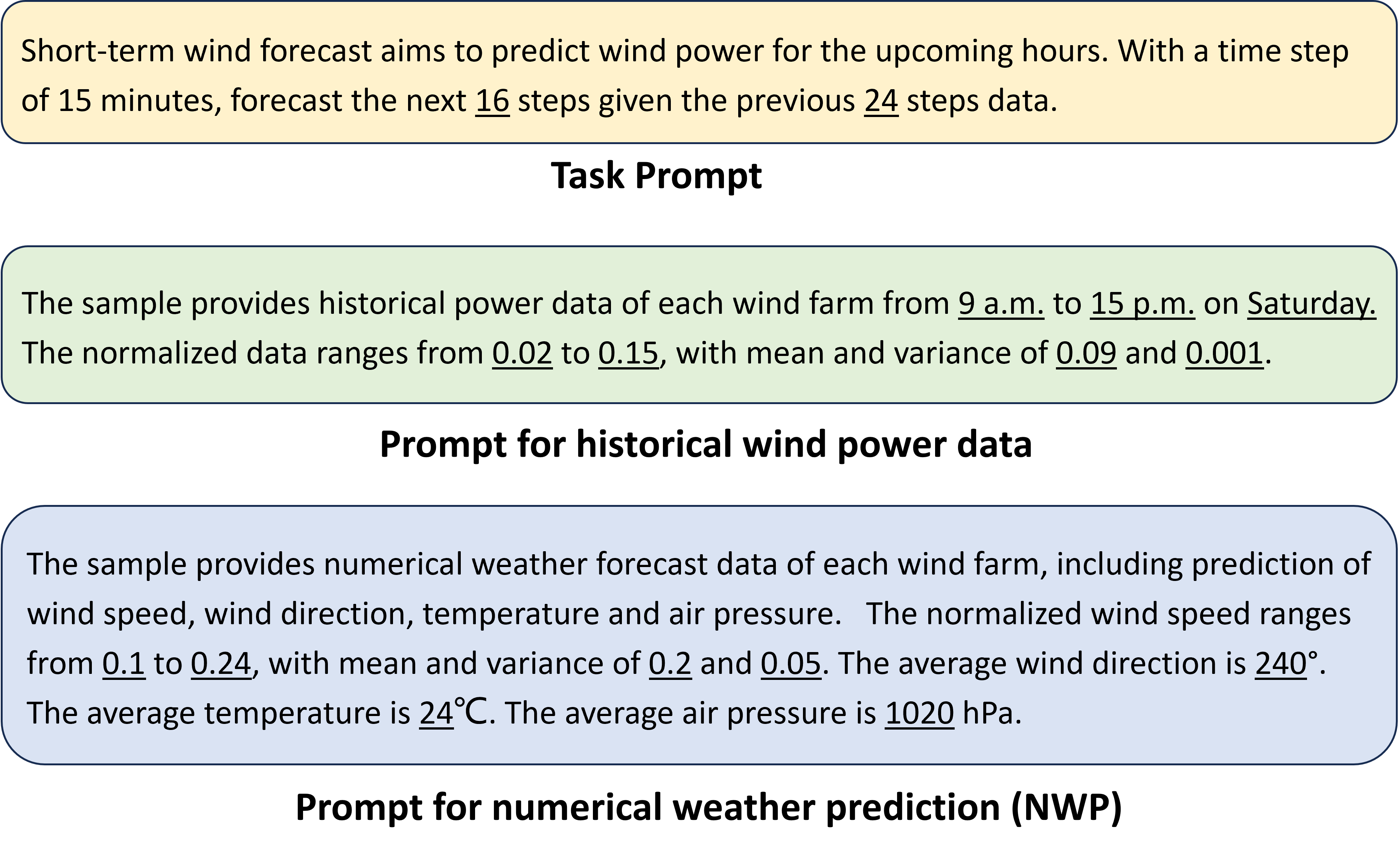}
		\caption{Prompt for wind forecast task}
		\label{fig:prompt}
	\end{figure}

	A well-crafted prompt prefix helps the LLMs quickly understand the downstream task and grasp the necessary information. Considering that an overly long prompt would cause the prefix to take up a significant portion, reducing the LLM's attention to the appended temporal numerical data, the prompt mainly provides a concise high-level description of the task and the input data characteristic, as illustrated in Fig. \ref{fig:prompt}. Given the LLM's limited ability to interpret mathematical expressions, mathematical symbols are avoided when designing the prompt. The underlined sections in the figure indicate interchangeable content, which is adjusted according to different parameter settings and different input temporal data in each sample.
	
	The prompt for the wind forecast task can be divided into three parts, namely the task prompt, the prompt for historical wind power data and the prompt for numerical weather prediction. The task prompt primarily outlines the target variables to be predicted, along with the time intervals and the lengths of the input and output data. The prompt for the historical wind power data can describe the time window of the historical wind power data, the data range, and the statistics such as the mean and variance. The prompt for the numerical weather prediction includes the wind speed, wind direction, and the statistical information of them.

	The data description varies according to samples and features, detailing the meaning of the data, the time span it covers, and relevant statistics such as range, mean, and variance. Notably, each feature in the time series data is concatenated with its corresponding prompt. As a result, for each sample, $C$ prompts are generated, each corresponding to a different feature. This information in the data description aids the LLMs in gaining a deeper understanding of the data.
	
	With explicit definitions, the prompt text can be customized for each sample. Assuming that the prompt describing each feature includes $L_s$ tokens, it is fed into the pre-trained LLMs embedder, resulting in embedded prompt data $\mathbf{S}_p \in \mathbb{R}^{C \times L_s \times d_{LLM}}$, where $d_{LLM}$ is the hidden dimension of the LLMs. 
	
	\subsection{Data Embedder}
	The data embedder maps the raw time series data into a high-dimensional space to be fed into the LLMs. There are two key issues to address: first, how to enable the LLMs to effectively comprehend and process long time series data while fully exploring potential temporal correlations; second, how to reasonably map the time series data to high dimensions, leveraging the pre-trained knowledge of the LLMs to enhance its understanding. To tackle these challenges, the data embedder in the proposed model, as shown in Figure 2, includes two submodules: the patcher and the semantic augmenter, which will be explained in detail below.
	
	\subsubsection{Patching of Time series data}
	While it is well-known that the attention mechanism is effective in capturing long-range dependencies in sequence data, it may overlook important local, short-term patterns when the sequence becomes too long, potentially reducing the accuracy of forecasting models. To address this issue, a practical solution is to segment long sequences into smaller patches, which are then input into the model simultaneously. This allows the model to retain global information while effectively focusing on local dependencies in the time series.
	
	To mitigate the impact of distribution shifts in time series data, instance normalization is applied. Specifically, for each sample, the time series is normalized across the time steps for each feature. After normalization, the time series is segmented into patches. Let the original time series be denoted as $\mathbf{X}_{1:\tau_h} \in \mathbb{R}^{C \times \tau_h}$. After segmentation, each patch has a length of $l_p$, and the step size for segmentation is $s$. The first patch consists of $\mathbf{X}_{1:l_p}$, the second patch is $\mathbf{X}_{(1+s):(l_p+s)}$, and the third patch is $\mathbf{X}_{(1+2s):(l_p+2s)}$, continuing until the end of the original sequence. In total, there are $P = \lceil(\tau_h - l_p)/s\rceil + 1$ patches, where $\lceil \cdot \rceil$ represents the ceiling function.
	
	Following this process, the input data $\mathbf{X} \in \mathbb{R}^{C \times \tau_h}$ is first normalized and then restructured into patch data $\mathbf{X}_p \in \mathbb{R}^{C \times P \times l_p}$, which is subsequently fed into the semantic augmenter.

	\subsubsection{Semantic Augmenter}
	
	The semantic augmenter aims to help pre-trained LLMs to better understand the contextual information within the patches. The length of each patch data is $l_p$, which is typically much smaller than the hidden dimension $d_{LLM}$ of the LLMs. Consequently, it is necessary to apply an appropriate embedding before feeding the data into the LLMs. At this stage, a key challenge is how to effectively utilize the LLMs' pre-trained knowledge to enhance their understanding of time series data. Unlike textual data, numerical data cannot be directly tokenized and embedded—just as a floating-point number cannot be interpreted through a traditional dictionary.
	
\begin{figure}[htbp]
		\centering
		\includegraphics[width=0.4\linewidth]{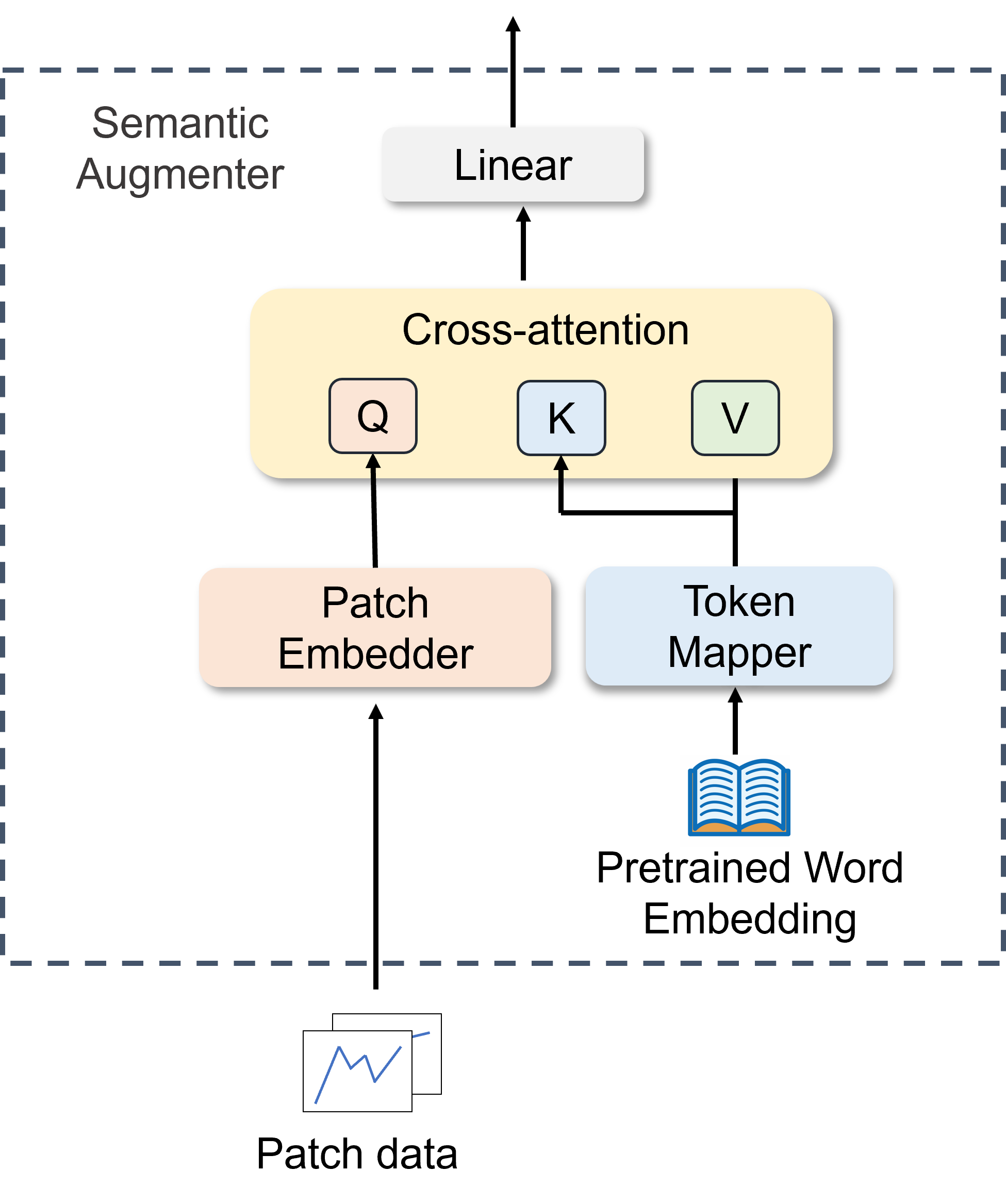}
		\caption{Attention-inspired Semantic Augmenter}
		\label{fig:Semantic_Augmenter}
	\end{figure}

	To address this, a semantic augmenter block is introduced to perform cross-attention between the patch data and the LLMs' pre-trained word embeddings, thereby uncovering potential relationships between the pre-trained knowledge and the time series data. This enables the time series data to be embedded in a meaningful way, allowing it to be aligned with and concatenated to prompt information before being processed by the LLMs. The structure of semantic augmenter is displayed in Fig. \ref{fig:Semantic_Augmenter}.
	
	The core mechanism of the semantic augmenter is to apply cross-attention between the patch data $\mathbf{X}_p$ and the pre-trained word embedding matrix $\mathbf{E}_0 \in {R}^{N_{v_0} \times d_{LLM}}$, where $N_{v_0}$ represents the vocabulary size, and $d_{LLM}$ is the hidden dimension of LLMs. 
	
	First, the patch data $\mathbf{X}_p$ is processed by a patch embedder module, which is a linear layer that transforms it from its original patch length $l_p$ to a hidden dimension $d_m$, aligning it with the cross-attention module. As a result, we obtain $\hat{\mathbf{X}}_p \in \mathbb{R}^{C \times P \times d_m}$. 
	
	Next, the pre-trained word embedding matrix $\mathbf{E}_0$ is mapped to a reduced vocabulary size through a linear layer, termed the Token Mapper. This reduces the embeddings to smaller, more representative categories, resulting in a mapped embedding matrix $\mathbf{E}_M \in \mathbb{R}^{N_{v_M} \times d_{LLMs}}$, where $N_{v_M} \ll N_{v_0}$. 
	
	We then perform multi-head cross-attention. For each head $i \in \{1, \dots, H\}$, where $H$ is the number of heads, the following matrices are acquired: the query matrix $\mathbf{Q}_i=\hat{\mathbf{X}}_p \mathbf{W}_i^Q$, the key matrix $\mathbf{K}_i={\mathbf{E}}_M \mathbf{W}_i^K$, and value matrix $\mathbf{V}_i={\mathbf{E}}_M \mathbf{W}_i^V$, where $\mathbf{W}_i^Q \in \mathbb{R}^{d_{m} \times d_H}, \mathbf{W}_i^K \in \mathbb{R}^{d_{LLMs} \times d_H}, \mathbf{W}i^V \in \mathbb{R}^{d_{LLMs} \times d_H}$, and $d_H = \lfloor d_m / H \rfloor$ is the dimensionality of each head (note that $\lfloor \cdot \rfloor$ denotes the floor function).
	
	The cross-attention for each head $i$ is computed as
	
	\begin{equation}
	\mathbf{Z}_i = \text{Softmax}\left( \frac{\mathbf{Q}_i \mathbf{K}_i^T}{\sqrt{d_H}} \right) \mathbf{V}_i
	\end{equation}
	
	By concatenating the outputs $\mathbf{Z}_i$ of all heads $i \in \{1, \dots, H\}$, we obtain the final output $\mathbf{Z} \in \mathbb{R}^{C \times P \times d_m}$. Then after passing $\mathbf{Z}$ through a linear layer, the augmented data $\mathbf{S}_d \in \mathbb{R}^{C \times P \times d_{LLMs}}$ is obtained, which can subsequently be used as part of the input to the LLMs.
	
	The above procedure effectively integrates contextual and semantic knowledge into the time-series representation. With the semantic augmenter, time-series data transforms from a mere sequence of numerical values into representations that capture both numerical and semantic contexts provided by the pre-trained language model. This augmentation enhances the LLMs' understanding of specific data and tasks, as all weight parameters within the semantic augmenter are trainable. This adaptability allows the embedding process to cater to different tasks and datasets, empowering the LLMs to perform more effectively in downstream forecasting tasks.
	
	\subsection{Fine-Tuned LLMs and Output Layer}
	
	By concatenating the embedded prompt data $\mathbf{S}_p$ and the embedded time series data $\mathbf{S}_d$, we obtain the input for the LLMs, denoted as $\mathbf{S}_c$. Although the pre-trained LLMs can directly process $\mathbf{S}_c$ and produce corresponding outputs, we aim to fine-tune the LLMs further to enhance prediction accuracy. To achieve this, we employ the well-known LoRA \cite{hu2021lora} technique for fine-tuning.
	
\begin{figure}[htbp]
		\centering
		\includegraphics[width=0.4\linewidth]{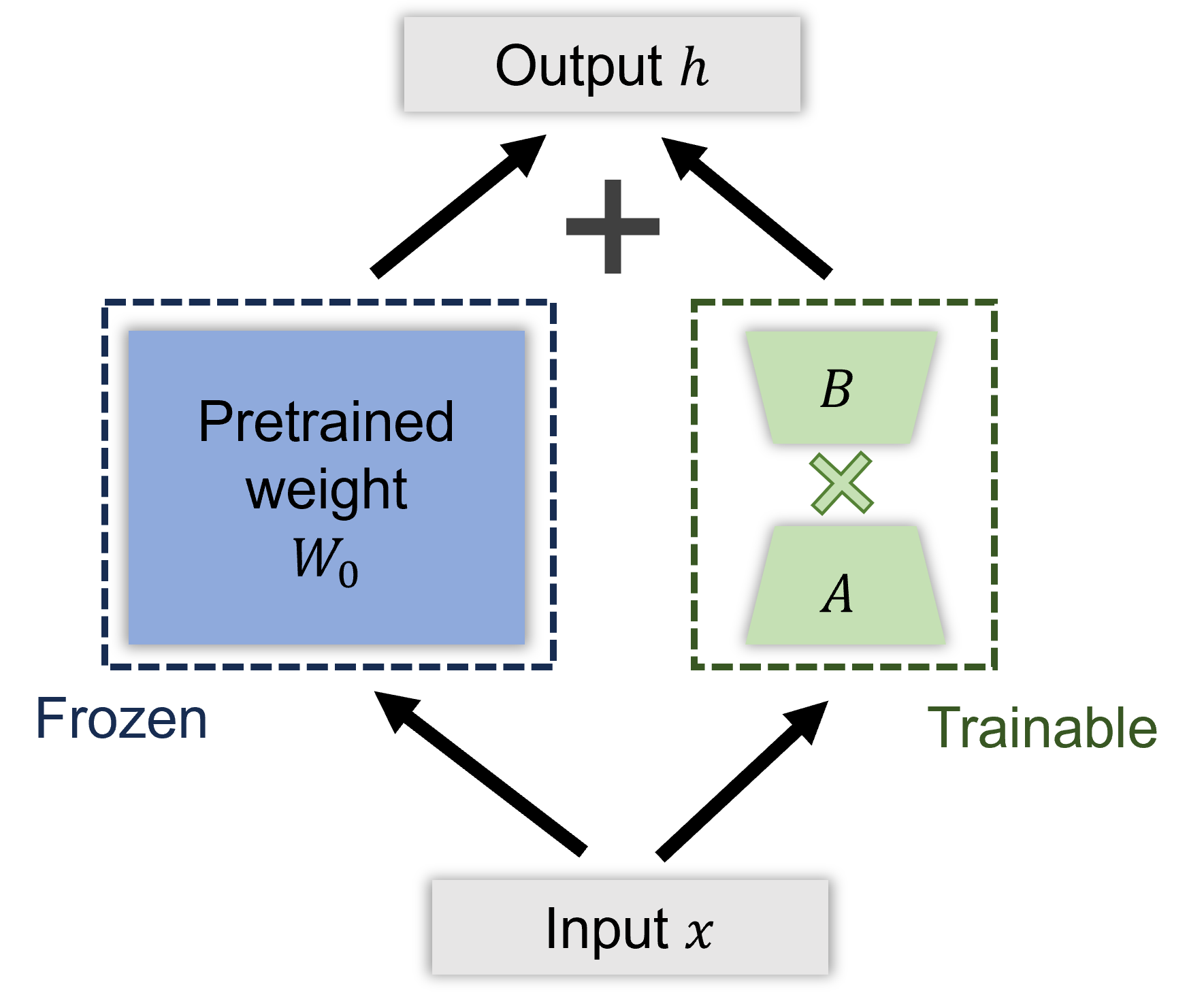}
		\caption{Illustration of LoRA}
		\label{fig:LoRA}
	\end{figure}

	Specifically, each layer of the LLMs contain high-dimensional parameters $W_0\in \mathbb{R}^{m \times n}$. Training full parameters might be challenging with limited samples. Instead, LoRA introduces a bypass branch at each layer, allowing the incorporation of a small number of trainable parameters $A\in \mathbb{R}^{r \times n}$ and $B\in \mathbb{R}^{m \times r}$ (where $r \ll min(m, n)$),  as illustrated in Fig. \ref{fig:LoRA}. These parameters $A$ and $B$ can be optimized using gradient descent based on the training samples, as their dimensions are significantly smaller than those of $W_0$. The final parameters $W$ are obtained by summing these with the frozen parameters $W_0$, expressed as 
	$W = W_0 + BA$.  This approach allows the LLMs to adapt effectively to downstream prediction tasks through the training of parameters $A$ and $B$.
	
	Subsequently, the output from the LLMs is processed through a linear layer to condense the information into the predicted time series and reshape it to match the shape of the forecast sequence, resulting in $\tilde{Y} \in \mathbb{R}^{C \times \tau_f}$. After denormalizing $\tilde{Y}$, the final forecast result $\hat{Y}$ is obtained.

	\section{Experiments}
	\subsection{Experimental setup}
	
	Experiments are conducted using historical wind power data from wind stations in Inner Mongolia, Yunnan, and Gansu, China. Each region contains five wind farms, with capacities ranging from 50 MW to 200 MW. Time series data is collected at 15-minute intervals. The input sequence length $\tau_h$ is set to 96, representing the past 24 hours of historical data, while the prediction length $\tau_f$ is 16, corresponding to the forecasted wind power output for the next 4 hours. The input features include historical wind power output and NWP features (including prediction of wind speed, air pressure, and temperature) at each wind power station. The dataset, comprising 17,546 samples collected from March 1, 2020, to August 31, 2020, is split into training, validation, and test sets in a 3:1:1 ratio. 
	
	Experiments are performed on an Ubuntu 18.04 system with four NVIDIA RTX 4090 GPUs equipped with 24GB of memory. The methods are implemented using PyTorch, with GPT-2 serving as the LLMs backbone for LLM-related methods (i.e., the proposed method and GPT4TS).

	\subsection{Evaluation metrics}
	Mean absolute error (MAE) and root mean square error (RMSE) are used to evaluate the forecast accuracy, i.e.,
	
	\begin{equation}
	MAE=\frac{1}{N_{test}} \sum_{i=1}^{N_{test}} \|\mathbf{Y}_i - \mathbf{\hat{Y}}_i\|
	\end{equation}
	\begin{equation}
	RMSE=\sqrt{\frac{1}{N_{test}} \sum_{i=1}^{N_{test}} \|\mathbf{Y}_i - \mathbf{\hat{Y}}_i\|^2}
	\end{equation}
	where $N_{test}$ is the number of test samples, $\mathbf{Y}_i$ is a vector of length $N_{test}$ which represents the actual values of wind power output for sample $i$, and $\mathbf{\hat{Y}}_i$ is also a vector of length $N_{test}$ which denotes the model's prediction. For each predicted time step, we can calculate the evaluation metrics separately.
	
	\subsection{Compared methods}
	
	The following methods are compared in this case study.
	\begin{itemize}
		\item \textbf{M2WLLM}: The proposed M2WLLM model is used for wind forecast. The weights of the Semantic Augmenter, LoRA, and the output linear layers are fine-tuned using the training set.
		\item \textbf{LSTM}\cite{LSTM}: The vanilla LSTM network is used for wind power forecast, with two hidden layers, and the dimension of features is 32 in each hidden layer.
		\item \textbf{TCN}\cite{TCN}: The Temporal Convolutional Network (TCN) utilizes 1D convolutions with increasing receptive fields, allowing it to efficiently model long-term dependencies in time series data.
		\item \textbf{Transformer}\cite{Transformer}: The vanilla transformer model includes an encoder and a decoder, each with eight self-attention blocks. 
		\item \textbf{Informer}\cite{Informer}: As one of the latest advancements in time series forecasting, the Informer model employs an attention mechanism specifically designed for efficient long sequence forecasting, leveraging a generative approach to capture temporal dependencies while reducing computational complexity.
		\item \textbf{Autoformer}\cite{Autoformer}: The Autoformer is a forecasting model that follows a residual and encoder-decoder structure. It excels at long-term time series predictions by decomposing data into trend and seasonal components and using an efficient Auto-Correlation mechanism to identify patterns.
		\item \textbf{Adaptive-GCN}\cite{yang2024centralized}: The approach integrates graph neural networks with NWP data and employs multi-modal learning to fuse historical wind power and NWP information. It also uses multi-task learning to predict power output for each individual wind farm within a cluster while improving efficiency. The model is designed to capture complex spatiotemporal relationships and nonlinear characteristics of wind power.
		\item \textbf{CNN-BiLSTM-Att}\cite{peng2023novel}: The predictive model proposed in the paper integrates the features of Convolutional Neural Networks (CNN) and Bidirectional Long Short-Term Memory Networks (Bi-LSTM) through a cross-attention mechanism. CNN extracts local features of wind power generation time series data through its convolutional layers, while Bi-LSTM extracts features of time series data through two LSTM networks in forward and backward directions. The cross-attention mechanism allows the model to establish connections between two different sets of features.
		\item \textbf{GPT4TS}\cite{GPT4TS}: The GPT4TS model adapts the capabilities of LLMs for time series forecasting, integrating temporal embeddings and attention mechanisms to effectively predict future values based on historical data. It uses GPT2 as the backbone. Unlike the proposed method, GPT4TS does not utilize prompt information or pre-trained word embeddings. 
		
	\end{itemize}
	
	The parameters of each model in the experiments are summarized in Table \ref{table:model_param}. In this table, $d_{model}$ represents the dimension of hidden states in the model, $n_{layer}$ denotes the number of layers in the model, and $n_{head}$ indicates the number of heads in the multi-head attention blocks. Additionally, $bs$ and $lr$ represent the batch size and learning rate, respectively. It is worth noting that in the proposed method, $d_{model}$ represents the hidden dimension of the GPT-2 model, which is 768. 
	
	\begin{table}[h]
		\centering
		\scriptsize
		\renewcommand{\arraystretch}{1.3}
		\caption{Hyperparameters of models}  \label{table:model_param}     
		\begin{tabular}{c|cccccc}
			\toprule
			& $d_{model}$  & $n_{layer}$ & $n_{head}$ & $bs$ & $lr$  \\
			\midrule
			LSTM & 32   & 4 & - & 64 & 0.05  \\
			TCN & 32   & 4 & - & 64 & 0.05  \\
			Transformer  & 128  & 4 & 8 & 64 & 0.05 \\
			Informer & 128  & 4 & 8 & 64 & 0.05  \\
			Autoformer & 128  & 4 & 8 & 64 & 0.05  \\
			Adaptive-GCN & 128 & 2 & - & 64 & 0.005 \\
			CNN-BiLSTM-Att & 128 & 2 & 8 & 64 & 0.005 \\
			GPT4TS & 768 & 8 & 8 & 32 & 0.0005 \\
			M2WLLM & 768  & 8 & 8 & 32 & 0.0005  \\
			\hline
		\end{tabular}
		
	\end{table}
	
	\subsection{Results}
	
	Table \ref{table:result_comp} presents the MAE and RMSE of different models on the three test sets for the 15-minute, 1-hour, 2-hour, and 4-hour prediction horizons, corresponding to the 1st, 4th, 8th, and 16th forecast steps, respectively. Bolded values indicate the best performance in each group. 
	
	\begin{table}[htp]
		\centering
		\scriptsize
		\renewcommand{\arraystretch}{1.1}
		\caption{Forecast results on the test set on different datasets} \label{table:result_comp}     
		\begin{tabular}{c|c|cc|cc|cc|cc}
			\toprule
			\multirow{2}{*}{Dataset} & \multirow{2}{*}{Model} & \multicolumn{2}{c|}{15 min (1 step)} & \multicolumn{2}{c|}{1 h (4 steps)} & \multicolumn{2}{c|}{2 h (8 steps)} & \multicolumn{2}{c}{4 h (16 steps)} \\
			\cline{3-10}
			& & MAE & RMSE & MAE & RMSE & MAE & RMSE & MAE & RMSE \\
			\midrule
			\multirow{9}{*}{\shortstack{Inner \\ Mongolia}}
			& LSTM\cite{LSTM} & 10.44 & 15.79 & 10.68 & 16.16 & 11.09 & 16.36 & 12.86 & 17.50 \\
			& TCN\cite{TCN} & 8.68 & 12.39 & 9.26 & 13.07 & 10.24 & 14.27 & 12.22 & 16.57 \\
			& Transformer\cite{Transformer} & 5.98 & 8.56 & 7.66 & 10.81 & 9.67 & 13.17 & 12.33 & 16.09 \\
			& Informer\cite{Informer} & 7.19 & 10.61 & 9.02 & 13.06 & 10.56 & 14.80 & 12.21 & 17.24 \\
			& Autoformer\cite{Autoformer} & 11.63 & 15.75 & 11.74 & 15.91 & 12.16 & 16.45 & 13.30 & 17.88 \\
			& Adaptive-GCN\cite{yang2024centralized} &5.02&7.23&7.19&10.33&9.36&13.28&11.46&15.95\\
			& CNN-BiLSTM-Att\cite{peng2023novel} &4.53 & 6.64 &7.03&10.27&9.16&12.89&11.50&15.76\\
			& GPT4TS\cite{GPT4TS} & 5.12 & 7.81 & 7.62 & 10.78 & 9.52 & 13.27 & 12.07 & 16.39 \\
                & M2WLLM & \textbf{3.36} & \textbf{5.24} & \textbf{6.59} & \textbf{9.57} & \textbf{8.76} & \textbf{12.21} & \textbf{11.37} & \textbf{15.26} \\
			\midrule
			\multirow{9}{*}{Gansu} 
			& LSTM\cite{LSTM} & 16.61 & 26.86 & 17.05 & 27.70 & 17.50 & 28.33 & 18.98 & 29.17 \\
			& TCN\cite{TCN} & 14.79 & 22.44 & 15.83 & 24.27 & 16.86 & 25.95 & 19.16 & 28.77 \\
			& Transformer\cite{Transformer} & 11.60 & 19.49 & 13.84 & 22.75 & 16.97 & 26.92 & 19.34 & 31.08 \\
			& Informer\cite{Informer} & 12.18 & 19.44 & 16.09 & 24.58 & 17.26 & 25.92 & 18.71 & 29.58 \\
			& Autoformer\cite{Autoformer} & 20.93 & 30.02 & 21.19 & 30.29 & 21.90 & 31.48 & 23.48 & 34.15 \\
			& Adaptive-GCN\cite{yang2024centralized} &9.55&14.65&13.98&22.06&17.27&27.01&19.01&29.37\\
			& CNN-BiLSTM-Att\cite{peng2023novel} &9.31&14.11&13.91&22.08&16.83&26.38&18.38&28.31\\
			& GPT4TS\cite{GPT4TS} & 8.75 & 13.48 & 13.69 & 20.39 & 16.70 & 24.36 & 19.55 & 28.00 \\
			& M2WLLM & \textbf{6.68} & \textbf{10.74} & \textbf{12.10} & \textbf{19.29} & \textbf{15.57} & \textbf{23.56} & \textbf{18.13} & \textbf{26.66} \\
			\midrule
			\multirow{9}{*}{Yunnan} 
			& LSTM\cite{LSTM} & 6.70 & 10.57 & 6.88 & 10.92 & 7.29 & 11.60 & 8.34 & 12.98 \\
			& TCN\cite{TCN} & 7.11 & 10.40 & 7.64 & 10.96 & 8.39 & 11.84 & 9.44 & 13.24 \\
			& Transformer\cite{Transformer} & 6.38 & 9.82 & 7.14 & 10.80 & 7.70 & 11.40 & 9.49 & 13.27 \\
			& Informer\cite{Informer} & 6.00 & 9.16 & 6.68 & 9.85 & 7.10 & 10.25 & 8.57 & 12.33 \\
			& Autoformer\cite{Autoformer} & 8.18 & 11.78 & 8.30 & 11.96 & 8.58 & 12.36  & 9.27 & 13.45 \\
			& Adaptive-GCN\cite{yang2024centralized} &3.79&5.97&5.39&8.28&6.67&10.07&8.20&12.19\\
			& CNN-BiLSTM-Att\cite{peng2023novel} &4.17&5.99&5.07&7.82&6.33&9.52&8.35&11.81\\
			& GPT4TS\cite{GPT4TS} & 3.59 & 5.88 & 5.33 & 8.43 & 6.57 & 10.15 & 8.28 & 12.45 \\
                & M2WLLM & \textbf{2.58} & \textbf{4.64} & \textbf{4.48} & \textbf{7.40} & \textbf{5.99} & \textbf{9.27} & \textbf{7.87} & \textbf{11.59}  \\
			\bottomrule
		\end{tabular}
	\end{table}
	
	It can be seen from Table \ref{table:result_comp} that the proposed M2WLLM model consistently outperforms all other methods across each dataset and prediction horizon. This comprehensive superiority demonstrates its effectiveness and robustness in diverse scenarios. In comparison, although GPT4TS is based on the same LLM backbone, its advantages over other methods are not as significant as those of the proposed approach. Notably, on the Inner Mongolia dataset, GPT4TS even lags behind Adaptive-GCN and CNN-BiLSTM-Att. This limitation arises because GPT4TS, despite fine-tuning the LLMs backbone, directly feeds time series patches into the LLMs without fully leveraging its pre-trained capability to understand textual sequences. In contrast, M2WLLM incorporates architectural improvements by introducing tailored prompts and reformulating time series data for historical power data and weather forecasts. This design maximizes the utilization of pre-trained word embeddings, resulting in significant accuracy gains.
	
	\begin{figure*}[!htbp]
		\centering
		\includegraphics[width=0.9\textwidth]{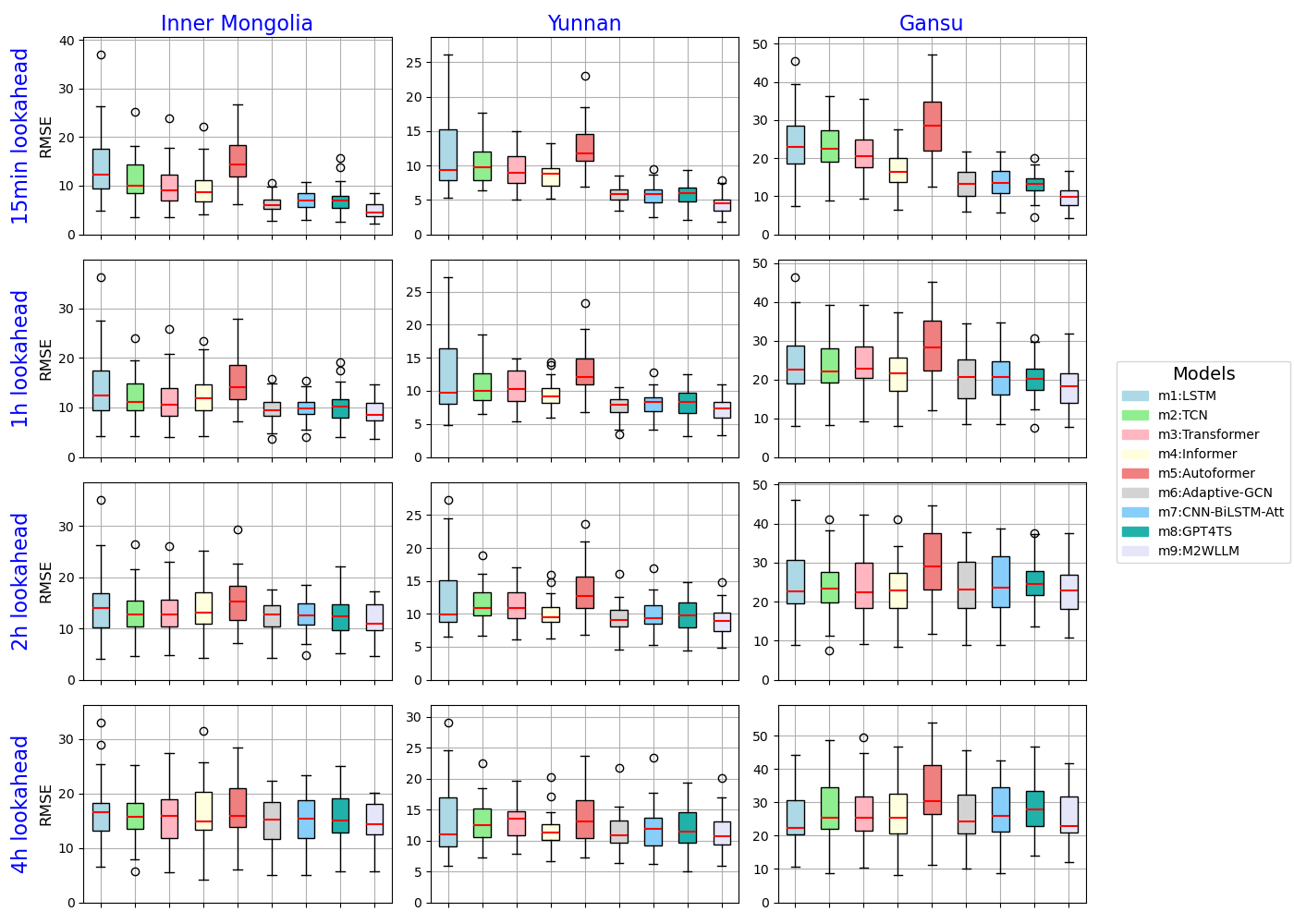}
		\caption{Distribution of daily RMSE on different datasets}
		\label{fig:boxplot}
	\end{figure*}
	
	Fig. \ref{fig:boxplot} illustrates the RMSE distribution for different forecast horizons across various datasets, visualized as box plots. Each data point corresponds to the daily RMSE (calculated over 96 consecutive time steps) from the test set. The median error, represented by the thick red line in the box plots, shows that M2WLLM consistently achieves lower overall prediction errors compared to other methods. Furthermore, the upper line of the box plots indicates that M2WLLM effectively avoids extreme prediction outliers, exhibiting smaller maximum errors than its counterparts. These results highlight the proposed method's superior accuracy and robustness across diverse datasets.
	
	\begin{figure*}[!htbp]
		\centering
		\includegraphics[width=0.8\textwidth]{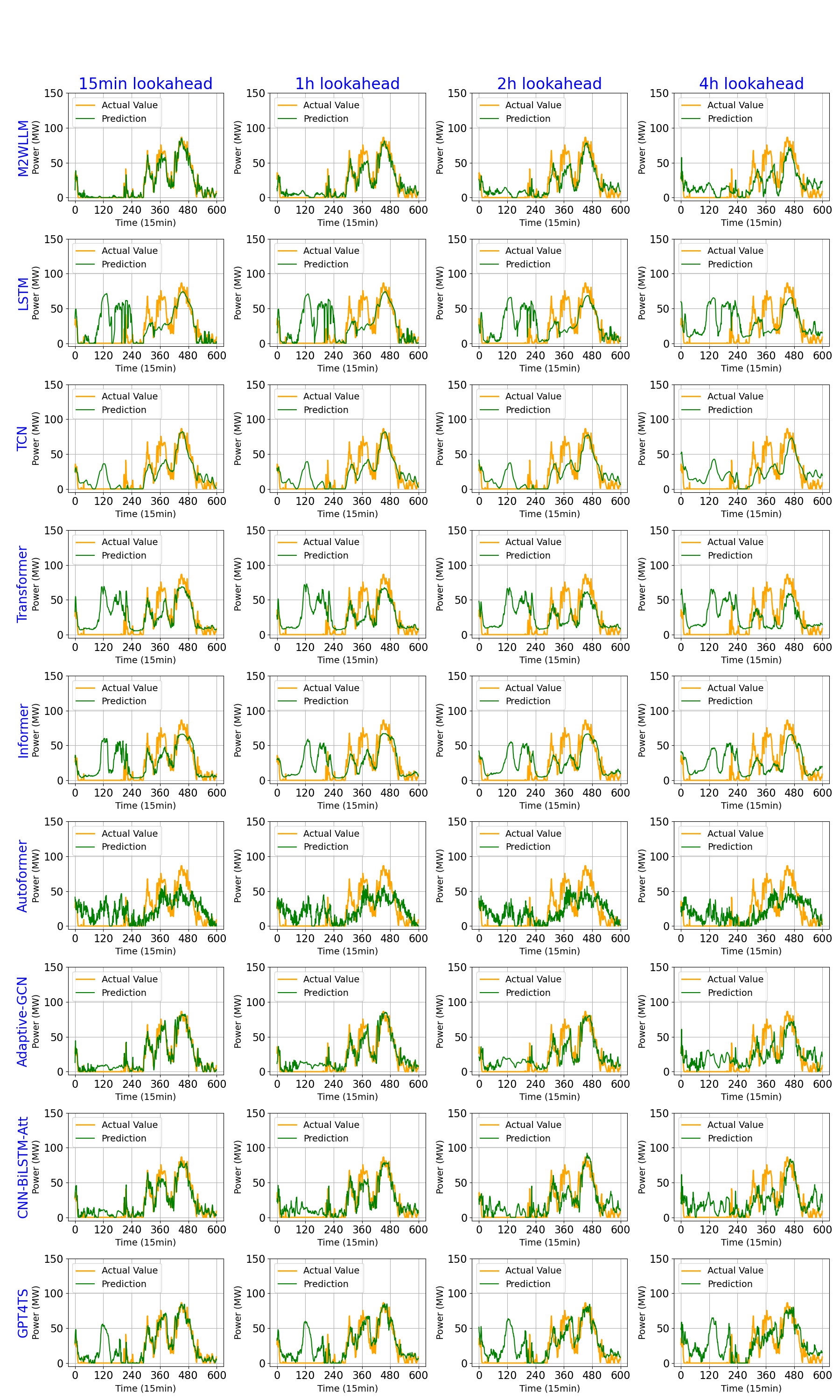}
		\caption{Forecast result of different methods}
		\label{fig:Forecast}
	\end{figure*}

	Fig. \ref{fig:Forecast} shows the actual wind power curve and the predictions of each method at different forecast horizons, based on 600 consecutive time periods randomly selected from the Inner Mongolia test set. The results demonstrate that the proposed M2WLLM model captures wind power fluctuations more accurately than other methods. Notably, during periods with no output, such as time steps 100–200, the proposed method provides reliable predictions, while other approaches tend to introduce extraneous fluctuations and fail to accurately depict the zero-output intervals. Compared to other methods, the proposed MW2LLM model can more effectively fit windless (zero output) periods. This is because the LLMs can fully understand the output patterns of windy and windless periods. Through semantic understanding and data semantic reconstruction, it more efficiently identifies the characteristics of wind power, thereby improving prediction accuracy.
	
	\begin{figure*}[!htbp]
		\centering
		\includegraphics[width=\textwidth]{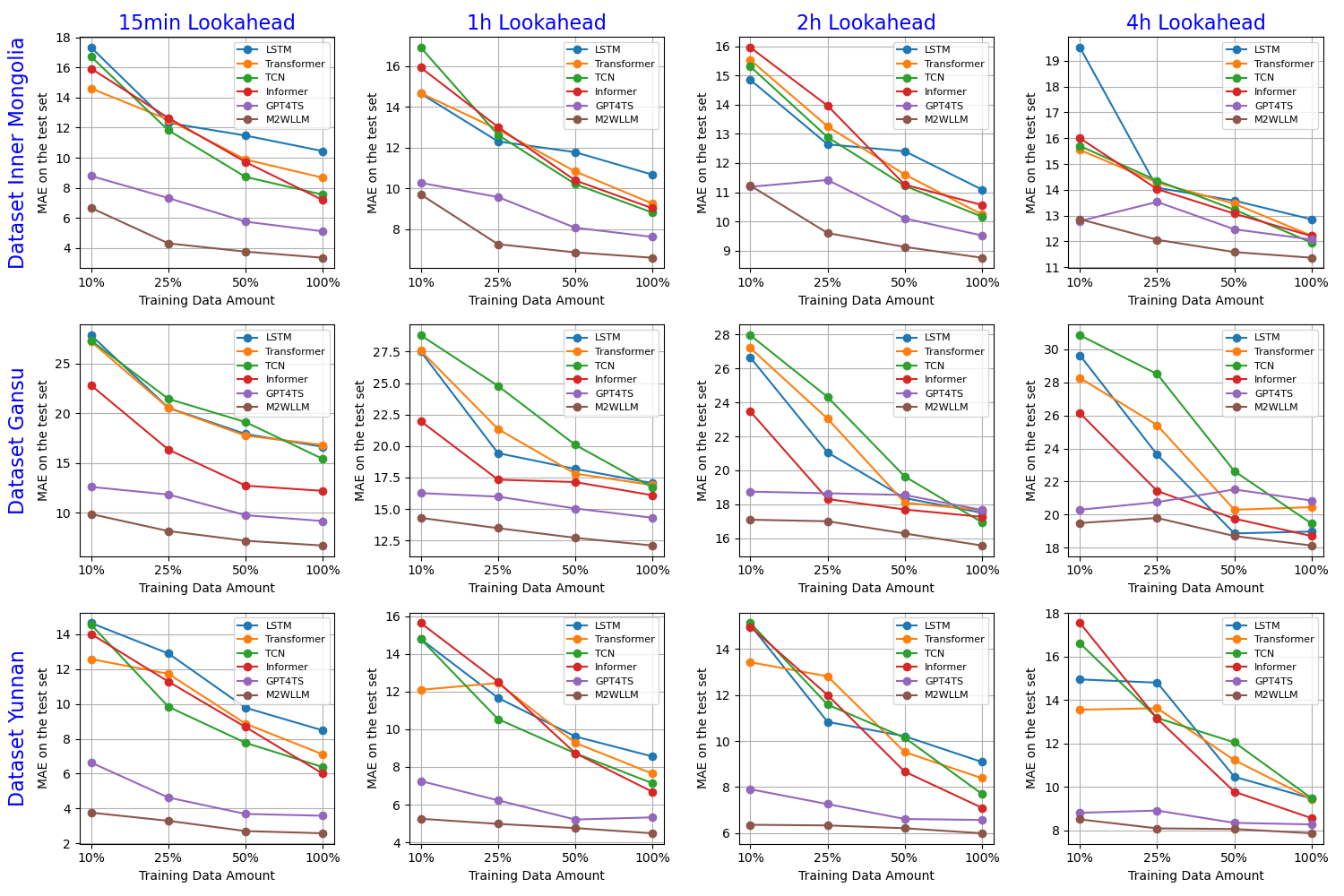}
		\caption{Model performance with different amounts of training data}
		\label{fig:MAE_Data_size}
	\end{figure*}
	
	Another point worth noting is the performance of the proposed method in few-shot learning. For newly established wind farms, historical data is often scarce, making it essential to evaluate the model's accuracy with limited samples. Fig. \ref{fig:MAE_Data_size} illustrates the forecasting performance of various methods across different datasets after training with varying amounts of data. To avoid an overly dense presentation of curves and legends, the figure selects six typical methods for comparison. As shown in  Fig. \ref{fig:MAE_Data_size}, the two LLM-based methods (M2WLLM and GPT4TS)  exhibit less performance fluctuation across different training data sizes compared to other methods In particular, the performance gap between using 10\% and 100\% of the training data is relatively small for these two methods, demonstrating the effectiveness of the LLM-based methods on few-shot learning. For example, in the Yunnan dataset's 15-minute lookahead forecast, training the Informer model with 10\% of the data versus 100\% results in a significant reduction in test set MAE, from 14.00 to 6.00. In contrast, training M2WLLM with 10\% and 100\% of the data yields a smaller difference, with MAE decreasing from 3.76 to 2.58. Similarly, for the 4-hour lookahead forecast on the Inner Mongolia dataset, training the LSTM model with 10\% of the data compared to 100\% leads to a significant drop in test set MAE, from 19.51 to 12.85. In contrast, the MAE for M2WLLM decreases only slightly, from 12.86 to 11.37, demonstrating its robustness. Furthermore, in many cases, M2WLLM’s performance with only 10\% of the training data surpasses that of other methods using 100\%. 
	For example, in the Yunnan dataset's 15-minute lookahead forecast, M2WLLM achieves a test set MAE of just 3.76 using only 10\% of the training data. In comparison, even with 100\% of the training data, Informer still has a higher test MAE of 6.00. Similarly, for the Gansu dataset's 4-hour lookahead forecast, M2WLLM achieves a test set MAE of 19.49 using only 10\% of the training data, which is comparable to the performance of other methods trained on 100\% of the data. This demonstrates the remarkable effectiveness of the proposed method in few-shot learning.

	\subsection{Ablation Study}
	
	To evaluate the value of the proposed model architecture, it is necessary to analyze performance changes by replacing its key components. Table \ref{table:abl_struct} presents the results after substituting three critical components of the proposed model. Specifically: 
	
	\begin{itemize}
		\item \textbf{w/o prompt}: This represents the prediction results without using a text prompt, achieved by replacing the prompt with an empty string.
		\item \textbf{w/o semantic augmenter}: This involves replacing the proposed semantic augmenter with a simple linear layer, bypassing pre-trained word embeddings, and directly applying linear embedding to the patch data.
		\item \textbf{w/o fine-tuning}: This denotes results obtained by keeping the LLMs backbone parameters fixed without any fine-tuning.
	\end{itemize}

	\begin{table*}[!htbp]
		\centering
		\scriptsize
		\renewcommand{\arraystretch}{1.3}
		\caption{Ablation studies on the model structure} \label{table:abl_struct}     
		\begin{tabular}{c|c|cc|cc|cc|cc}
			\toprule
			\multirow{2}{*}{Dataset} & \multirow{2}{*}{Model} & \multicolumn{2}{c|}{Step 1} & \multicolumn{2}{c|}{Step 4} & \multicolumn{2}{c|}{Step 8} & \multicolumn{2}{c}{Step 16} \\
			\cline{3-10}
			& & MAE & RMSE & MAE & RMSE & MAE & RMSE & MAE & RMSE \\
			\midrule 
			\multirow{4}{*}{\shortstack{Inner \\ Mongolia}} 
			& M2WLLM & \textbf{3.36} & \textbf{5.24} & \textbf{6.59} & \textbf{9.57} & \textbf{8.76} & \textbf{12.21} & \textbf{11.37} & \textbf{15.26} \\
			& w/o prompt & 4.60 & 6.76 & 7.04 & 10.06 & 9.03 & 12.50 & 11.49 & 15.40 \\
			& w/o semantic augmenter & 5.28 & 7.64 & 7.45 & 10.51 & 9.19 & 12.61 & 11.75 & 15.50 \\
			& w/o fine-tuning & 5.10 & 7.36 & 7.21 & 10.30 & 8.94 & 12.49 & 11.59 & 15.32 \\
			\midrule
			\multirow{4}{*}{Gansu} 
			& M2WLLM & \textbf{6.68} & \textbf{10.74} & \textbf{12.10} & \textbf{19.29} & \textbf{15.57} & \textbf{23.56} & \textbf{18.13} & \textbf{26.66} \\
			& w/o prompt & 9.83 & 14.98 & 13.81 & 20.82 & 16.67 & 24.27 & 19.21 & 27.27 \\
			& w/o semantic augmenter & 9.23 & 14.47 & 13.41 & 20.59 & 16.23 & 23.95 & 18.86 & 27.03 \\
			& w/o fine-tuning & 10.72 & 16.30 & 14.31 & 21.48 & 16.97 & 24.57 & 19.66 & 27.88 \\
			\midrule
			\multirow{4}{*}{Yunnan} 
			& M2WLLM & \textbf{2.58} & \textbf{4.64} & \textbf{4.48} & \textbf{7.40} & \textbf{5.99} & \textbf{9.27} & \textbf{7.87} & \textbf{11.59} \\
			& w/o prompt & 2.74 & 4.86 & 4.67 & 7.50 & 6.13 & 9.32 & 7.98 & 11.78 \\
			& w/o semantic augmenter & 3.75 & 5.97 & 5.26 & 8.07 & 6.54 & 9.72 & 8.18 & 11.78 \\
			& w/o fine-tuning & 4.38 & 6.70 & 5.75 & 8.59 & 6.91 & 10.12 & 8.43 & 12.13 \\
			\bottomrule
		\end{tabular}
	\end{table*}
	
	From the results in Table \ref{table:abl_struct}, it is evident that replacing any of the key modules leads to an increase in prediction error across all datasets, indicating that each module contributes to the model's performance. The greater the increase in error, the more critical the module is to the model. Overall, results across the three datasets suggest that the semantic augmenter and fine-tuning have a more significant impact. The semantic augmenter translates time series data into a form that is more interpretable for the LLMs, while the fine-tuning module adjusts a small number of parameters to quickly adapt the LLM to the specific wind power forecasting task. At the same time, removing the prompt module also results in an increase in forecast error, demonstrating its value in enhancing the LLMs' understanding. In summary, each module plays a unique and essential role in the model's effectiveness.
	
	\begin{table*}[!htp]
		\centering
		\scriptsize
		\renewcommand{\arraystretch}{1.3}
		\caption{Ablation studies on NWP information} \label{table:abl_NWP}     
		\begin{tabular}{c|c|cc|cc|cc|cc}
			\toprule
			\multirow{2}{*}{Dataset} & \multirow{2}{*}{Model} & \multicolumn{2}{c|}{Step 1} & \multicolumn{2}{c|}{Step 4} & \multicolumn{2}{c|}{Step 8} & \multicolumn{2}{c}{Step 16} \\
			\cline{3-10}
			& & MAE & RMSE & MAE & RMSE & MAE & RMSE & MAE & RMSE \\
			\midrule 
			\multirow{2}{*}{\shortstack{Inner \\ Mongolia}} 
			& with NWP information & \textbf{3.36} & \textbf{5.24} & \textbf{6.59} & \textbf{9.57} & \textbf{8.76} & \textbf{12.21} & \textbf{11.37} & \textbf{15.26}
			\\
			& without NWP information & 3.65 & 5.72 & 6.94 & 10.55 & 9.36 & 13.53 & 12.25 & 17.09 \\
			\midrule
			\multirow{2}{*}{Gansu} 
			& with NWP information & \textbf{6.68} & \textbf{10.74} & \textbf{12.10} & \textbf{19.29} & \textbf{15.57} & \textbf{23.56} & \textbf{18.13} & \textbf{26.66}
			\\
			& without NWP information & 6.76 & 10.75 & 14.61 & 22.13 & 17.63 & 27.54 & 23.25 & 33.38 \\
			\midrule
			\multirow{2}{*}{Yunnan} 
			& with NWP information & \textbf{2.58} & \textbf{4.64} & \textbf{4.48} & \textbf{7.40} & \textbf{5.99} & \textbf{9.27} & \textbf{7.87} & \textbf{11.59} \\
			& without NWP information & 3.12 & 5.42 & 5.24 & 8.25 & 6.62 & 10.04 & 8.58 & 12.50 \\
			\bottomrule
		\end{tabular}
	\end{table*}      
	
	Table \ref{table:abl_NWP} presents the wind power prediction errors with and without NWP information. The results show that incorporating NWP information consistently reduces prediction errors under all conditions, with its impact becoming more pronounced as the forecast horizon increases. This is because, for longer forecast horizons, historical data provides limited insight into distant future conditions, making NWP information critical for forming a clearer understanding of future trends. The proposed model effectively interprets and leverages NWP information, thereby enhancing its overall performance.
	
	Table \ref{table:abl_GPT_layer} summarizes the prediction performance of the proposed model when using different numbers of GPT-2 layers as the LLMs backbone. GPT-2 consists of 12 layers in total, and the default experiments in this study use the first 8 layers. The comparison in Table \ref{table:abl_GPT_layer} further evaluates the results with 2, 4, 8, and 12 layers.
	
	\begin{table*}[!htp]
		\centering
		\scriptsize
		\renewcommand{\arraystretch}{1.3}
		\caption{Ablation studies on different GPT layers} \label{table:abl_GPT_layer}     
		\begin{tabular}{c|c|cc|cc|cc|cc}
			\toprule
			\multirow{2}{*}{Dataset} & \multirow{2}{*}{Model} & \multicolumn{2}{c|}{Step 1} & \multicolumn{2}{c|}{Step 4} & \multicolumn{2}{c|}{Step 8} & \multicolumn{2}{c}{Step 16} \\
			\cline{3-10}
			& & MAE & RMSE & MAE & RMSE & MAE & RMSE & MAE & RMSE \\
			\midrule 
			\multirow{4}{*}{\shortstack{Inner \\ Mongolia}} 
			& layer 2 & 3.41 & 5.17 & 6.41 & 9.43 & 8.80 & 12.23 & 11.33 & 15.16\\
			& layer 4 & 3.33 & \textbf{5.16} & \textbf{6.29} & 9.45 & \textbf{8.52} & \textbf{12.16} & \textbf{11.03} & \textbf{15.10} \\
			& layer 8 & 3.36 & 5.24 & 6.59 & 9.57 & 8.76 & 12.21 & 11.37 & 15.26 \\
			& layer 12 & \textbf{3.32} & 5.19  & 6.40 & \textbf{9.39} & 8.74 & 12.18 & 11.26 & 15.12 \\
			\midrule
			\multirow{4}{*}{Gansu} 
			& layer 2 & 7.69 & 12.27 & 12.56 & 19.81 & 16.19 & 23.88 & 18.87 & 27.09\\
			& layer 4 & 7.16 & 11.38 & 12.40 & 19.46 & 15.59 & \textbf{23.37} & 18.24 & \textbf{26.26} \\
			& layer 8 & 6.68 & 10.74 & 12.10 & 19.29 & 15.57 & 23.56 & 18.13 & 26.66 \\
			& layer 12 & \textbf{6.42} & \textbf{10.65} & \textbf{11.88} & \textbf{19.23} & \textbf{15.27} & 23.47 & \textbf{17.62} & 26.46 \\
			\midrule
			\multirow{4}{*}{Yunnan} 
			& layer 2 & 2.64 & 4.69 & 4.53 & \textbf{7.37} & 6.13 & 9.36 & 7.96 & 11.67 \\
			& layer 4 & 2.55 & \textbf{4.58} & 4.54 & 7.38 & 6.11 & 9.30 & 8.05 & 11.79 \\
			& layer 8 & 2.58 & 4.64 & \textbf{4.48} & 7.40 & \textbf{5.99} & \textbf{9.27} & 7.87 & 11.59\\
			& layer 12 & \textbf{2.46} & 4.60 & 4.55 & 7.44 & 6.03 & 9.30 & \textbf{7.83} & \textbf{11.57} \\
			\bottomrule
		\end{tabular}
	\end{table*}     
	
	The results in Table \ref{table:abl_GPT_layer} show a general trend of improving forecast accuracy as the number of LLMs backbone layers increases. However, the performance improvement becomes negligible beyond 4 layers, indicating that at this point, the LLMs already possess sufficient capacity to handle the wind power forecasting task. This suggests that the effectiveness of the proposed model relies more on its architecture rather than the sheer parameter size of the LLMs. As a result, the computational requirements for training the model are relatively modest, making it more accessible and resource-friendly.
	
	\begin{figure*}[!htp]
		\centering
		\includegraphics[width=0.8\textwidth]{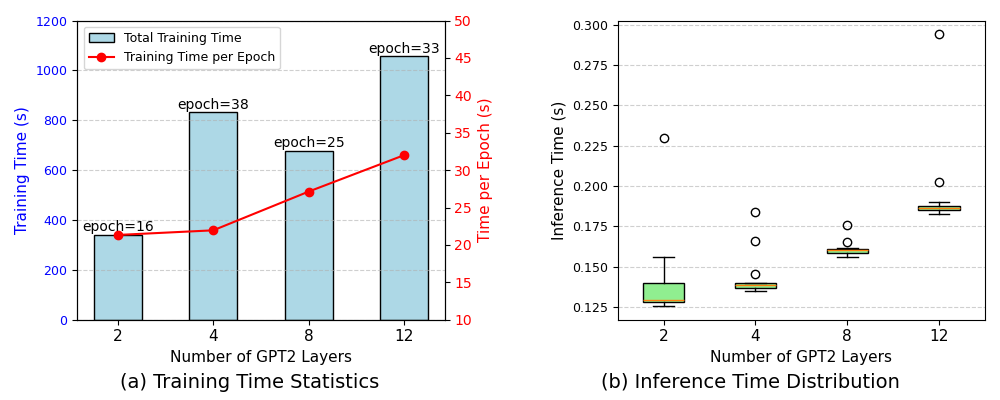}
		\caption{Training and inference time with different LLMs layers}
		\label{fig:Training_time}
	\end{figure*}
	
	Figure \ref{fig:Training_time} shows the training and inference times of the proposed model on dataset Gansu using different numbers of LLMs backbone layers. Training time is reported as both the total duration for all epochs and the average duration per epoch. Inference time refers to the time required to predict wind power for each batch in the test set, using a batch size of 32. The results indicate that both the average training time per epoch and the inference time increase with the number of LLMs layers, yet the overall efficiency remains high. For a training set with over 10,000 samples, each training epoch completes in within one minute, with the total training time within 30 minutes. Inference time is less than 0.5 seconds per batch, demonstrating the model's capability for efficient predictions.

	\section{Conclusion}
	
	The present study introduces M2WLLM, a novel approach to multi-modal multi-task ultra-short-term wind power forecasting that leverages the advanced capabilities of Large Language Models (LLMs). The method is verified in the wind power dataset from Inner Mongolia, Gansu, and Yunnan provinces. The main conclusions can be reached as follows:
    
     (1) M2WLLM demonstrates a significant enhancement in prediction accuracy over existing deep learning methods. Our experimental results, based on comprehensive datasets from wind farms in China, consistently show that M2WLLM outperforms other state-of-the-art forecasting models, including the LLM-based GPT4TS, across different prediction horizons. 
	
    (2) M2WLLM allows for an integrated fusion of textual prompts and numerical data, capitalizing on the pre-trained knowledge of LLMs to extract hidden temporal features effectively. The ablation studies confirm the importance of each component of the model, highlighting the synergistic effect of prompt embedding, semantic augmentation, and fine-tuning in achieving superior forecasting performance.
	
    (3) M2WLLM also exhibits robust performance even in few-shot learning scenarios, which is particularly relevant for newly established wind farms with limited historical data. This robustness, coupled with its high accuracy, positions M2WLLM as a valuable tool for the renewable energy industry, offering practical insights for grid operators and energy managers tasked with integrating intermittent wind power into the power grid.
	
    Although the paper mentions that M2WLLM has advantages in predictive accuracy, the current choice of base model is GPT-2 and other traditional large language base models. While these models are relatively powerful, they are somewhat lacking in model specificity, which may lead to some waste of computing resources. Therefore, in the future, we will further explore the training of base models suitable for wind farm power prediction using new energy and time series data, and attempt to extend the prediction methods to photovoltaic prediction and short-term as well as medium to long-term forecasting scenarios.

	\bibliographystyle{elsarticle-num} 
	\bibliography{refs_LLM}

\begin{thebibliography}{10}
\expandafter\ifx\csname url\endcsname\relax
  \def\url#1{\texttt{#1}}\fi
\expandafter\ifx\csname urlprefix\endcsname\relax\def\urlprefix{URL }\fi
\expandafter\ifx\csname href\endcsname\relax
  \def\href#1#2{#2} \def\path#1{#1}\fi

\bibitem{xue2015review}
Y.~Xue, Y.~Chen, J.~Zhao, X.~Liu, K.~Li, W.~Qiu, Y.~Gang, A review on
  short-term and ultra-short-term wind power prediction, Dianli Xitong
  Zidonghua/Automation of Electric Power Systems 39~(6) (2015) 1--12.

\bibitem{wang2024high}
S.~Wang, J.~Shi, W.~Yang, Q.~Yin, High and low frequency wind power prediction
  based on transformer and bigru-attention, Energy 288 (2024) 129753.

\bibitem{hong2020energy}
T.~Hong, P.~Pinson, Y.~Wang, R.~Weron, D.~Yang, H.~Zareipour, Energy
  forecasting: A review and outlook, IEEE Open Access Journal of Power and
  Energy 7 (2020) 376--388.

\bibitem{ARIMA_NP}
Q.~Han, F.~Meng, T.~Hu, F.~Chu, Non-parametric hybrid models for wind speed
  forecasting, Energy Conversion and Management 148 (2017) 554--568.

\bibitem{messner2019online}
J.~W. Messner, P.~Pinson, Online adaptive lasso estimation in vector
  autoregressive models for high dimensional wind power forecasting,
  International Journal of Forecasting 35~(4) (2019) 1485--1498.

\bibitem{XGBoost_BO}
X.~Xiong, X.~Guo, P.~Zeng, R.~Zou, X.~Wang, A short-term wind power forecast
  method via xgboost hyper-parameters optimization, Frontiers in energy
  research 10 (2022) 905155.

\bibitem{hu2021hybrid}
S.~Hu, Y.~Xiang, H.~Zhang, S.~Xie, J.~Li, C.~Gu, W.~Sun, J.~Liu, Hybrid
  forecasting method for wind power integrating spatial correlation and
  corrected numerical weather prediction, Applied Energy 293 (2021) 116951.

\bibitem{peng2023short}
X.~Peng, Z.~Yang, Y.~Li, B.~Wang, J.~Che, Short-term wind power prediction
  based on stacked denoised auto-encoder deep learning and multi-level transfer
  learning, Wind Energy 26~(10) (2023) 1066--1081.

\bibitem{LSTM_EMD_kNN}
K.~Sareen, B.~K. Panigrahi, T.~Shikhola, R.~Sharma, An imputation and
  decomposition algorithms based integrated approach with bidirectional lstm
  neural network for wind speed prediction, Energy 278 (2023) 127799.

\bibitem{LSTM_WT_CSA}
G.~Memarzadeh, F.~Keynia, A new short-term wind speed forecasting method based
  on fine-tuned lstm neural network and optimal input sets, Energy Conversion
  and Management 213 (2020) 112824.

\bibitem{LSTM_CNN_1}
C.~Yu, Y.~Li, L.~Zhao, Q.~Chen, Y.~Xun, A novel time-frequency recurrent
  network and its advanced version for short-term wind speed predictions,
  Energy 262 (2023) 125556.

\bibitem{ConvLSTM_2}
S.-X. Lv, L.~Wang, Multivariate wind speed forecasting based on multi-objective
  feature selection approach and hybrid deep learning model, Energy 263 (2023)
  126100.

\bibitem{Attention_2}
H.~Wu, K.~Meng, D.~Fan, Z.~Zhang, Q.~Liu, Multistep short-term wind speed
  forecasting using transformer, Energy 261 (2022) 125231.

\bibitem{Attention_1}
H.~Zhang, J.~Yan, Y.~Liu, Y.~Gao, S.~Han, L.~Li, Multi-source and temporal
  attention network for probabilistic wind power prediction, IEEE Transactions
  on Sustainable Energy 12~(4) (2021) 2205--2218.

\bibitem{LSTM_Attention_1}
X.~Liu, J.~Zhou, Short-term wind power forecasting based on
  multivariate/multi-step lstm with temporal feature attention mechanism,
  Applied Soft Computing 150 (2024) 111050.

\bibitem{LSTM_Attention_2}
Z.~Ma, G.~Mei, A hybrid attention-based deep learning approach for wind power
  prediction, Applied Energy 323 (2022) 119608.

\bibitem{peng2023novel}
C.~Peng, Y.~Zhang, B.~Zhang, D.~Song, Y.~Lyu, A.~Tsoi, A novel ultra-short-term
  wind power prediction method based on xa mechanism, Applied Energy 351 (2023)
  121905.

\bibitem{fan2020m2gsnet}
H.~Fan, X.~Zhang, S.~Mei, K.~Chen, X.~Chen, M2gsnet: Multi-modal multi-task
  graph spatiotemporal network for ultra-short-term wind farm cluster power
  prediction, Applied Sciences 10~(21) (2020) 7915.

\bibitem{liang2024wpformer}
X.~Liang, Q.~Gu, X.~You, Wpformer: A spatial-temporal graph transformer with
  auto-correlation for wind power forecasting, IEEE Transactions on Sustainable
  Energy (2024).

\bibitem{ye2022novel}
L.~Ye, Y.~Li, M.~Pei, Y.~Zhao, Z.~Li, P.~Lu, A novel integrated method for
  short-term wind power forecasting based on fluctuation clustering and history
  matching, Applied Energy 327 (2022) 120131.

\bibitem{von2021online}
L.~Von~Krannichfeldt, Y.~Wang, T.~Zufferey, G.~Hug, Online ensemble approach
  for probabilistic wind power forecasting, IEEE Transactions on Sustainable
  Energy 13~(2) (2021) 1221--1233.

\bibitem{li2023adaptive}
M.~Li, M.~Yang, Y.~Yu, M.~Shahidehpour, F.~Wen, Adaptive weighted combination
  approach for wind power forecast based on deep deterministic policy gradient
  method, IEEE Transactions on Power Systems (2023).

\bibitem{liu2023spatio}
X.~Liu, Y.~Zhang, Z.~Zhen, F.~Xu, F.~Wang, Z.~Mi, Spatio-temporal graph neural
  network and pattern prediction based ultra-short-term power forecasting of
  wind farm cluster, IEEE Transactions on Industry Applications (2023).

\bibitem{xu2023adaptive}
H.~Xu, Y.~Zhang, Z.~Zhen, F.~Xu, F.~Wang, Adaptive feature selection and gcn
  with optimal graph structure-based ultra-short-term wind farm cluster power
  forecasting method, IEEE Transactions on Industry Applications (2023).

\bibitem{zhao2024interpretable}
Y.~Zhao, H.~Liao, S.~Pan, Y.~Zhao, Interpretable multi-graph convolution
  network integrating spatio-temporal attention and dynamic combination for
  wind power forecasting, Expert Systems with Applications 255 (2024) 124766.

\bibitem{yang2024short}
M.~Yang, C.~Ju, Y.~Huang, Y.~Guo, M.~Jia, Short-term power forecasting of wind
  farm cluster based on global information adaptive perceptual graph
  convolution network, IEEE Transactions on Sustainable Energy (2024).

\bibitem{yang2024centralized}
M.~Yang, D.~Wang, W.~Zhang, X.~Yv, A centralized power prediction method for
  large-scale wind power clusters based on dynamic graph neural network, Energy
  310 (2024) 133210.

\bibitem{GPT4TS}
T.~Zhou, P.~Niu, L.~Sun, R.~Jin, et~al., One fits all: Power general time
  series analysis by pretrained {LM}, Advances in neural information processing
  systems 36 (2023) 43322--43355.

\bibitem{ansari2024chronos}
A.~F. Ansari, L.~Stella, C.~Turkmen, X.~Zhang, P.~Mercado, H.~Shen, O.~Shchur,
  S.~S. Rangapuram, S.~P. Arango, S.~Kapoor, et~al., Chronos: Learning the
  language of time series, arXiv preprint arXiv:2403.07815 (2024).

\bibitem{gruver2024large}
N.~Gruver, M.~Finzi, S.~Qiu, A.~G. Wilson, Large language models are zero-shot
  time series forecasters, Advances in Neural Information Processing Systems 36
  (2024).

\bibitem{jin2023time}
M.~Jin, S.~Wang, L.~Ma, Z.~Chu, J.~Y. Zhang, X.~Shi, P.-Y. Chen, Y.~Liang,
  Y.-F. Li, S.~Pan, Q.~Wen, Time-{LLM}: Time series forecasting by
  reprogramming large language models, in: The Twelfth International Conference
  on Learning Representations, 2024.

\bibitem{li2024urbangpt}
Z.~Li, L.~Xia, J.~Tang, Y.~Xu, L.~Shi, L.~Xia, D.~Yin, C.~Huang, Urbangpt:
  Spatio-temporal large language models, in: Proceedings of the 30th ACM SIGKDD
  Conference on Knowledge Discovery and Data Mining, 2024, pp. 5351--5362.

\bibitem{yu2023temporal}
X.~Yu, Z.~Chen, Y.~Ling, S.~Dong, Z.~Liu, Y.~Lu, Temporal data meets
  llm--explainable financial time series forecasting, arXiv preprint
  arXiv:2306.11025 (2023).

\bibitem{BERT4ST}
Z.~Lai, T.~Wu, X.~Fei, Q.~Ling, Bert4st:: Fine-tuning pre-trained large
  language model for wind power forecasting, Energy Conversion and Management
  307 (2024) 118331.

\bibitem{STELLM}
T.~Wu, Q.~Ling, Stellm: Spatio-temporal enhanced pre-trained large language
  model for wind speed forecasting, Applied Energy 375 (2024) 124034.

\bibitem{hu2021lora}
E.~J. Hu, Y.~Shen, P.~Wallis, Z.~Allen-Zhu, Y.~Li, S.~Wang, L.~Wang, W.~Chen,
  Lora: Low-rank adaptation of large language models, arXiv preprint
  arXiv:2106.09685 (2021).

\bibitem{LSTM}
S.~Hochreiter, Long short-term memory, Neural Computation MIT-Press (1997).

\bibitem{TCN}
C.~Lea, M.~D. Flynn, R.~Vidal, A.~Reiter, G.~D. Hager, Temporal convolutional
  networks for action segmentation and detection, in: proceedings of the IEEE
  Conference on Computer Vision and Pattern Recognition, 2017, pp. 156--165.

\bibitem{Transformer}
A.~Vaswani, Attention is all you need, Advances in Neural Information
  Processing Systems (2017).

\bibitem{Informer}
H.~Zhou, S.~Zhang, J.~Peng, S.~Zhang, J.~Li, H.~Xiong, W.~Zhang, Informer:
  Beyond efficient transformer for long sequence time-series forecasting, in:
  Proceedings of the AAAI conference on artificial intelligence, Vol.~35, 2021,
  pp. 11106--11115.

\bibitem{Autoformer}
H.~Wu, J.~Xu, J.~Wang, M.~Long, Autoformer: Decomposition transformers with
  auto-correlation for long-term series forecasting, Advances in neural
  information processing systems 34 (2021) 22419--22430.

\end{thebibliography}
	
\end{document}